\documentclass[conference]{IEEEtran}
\usepackage{times}

\usepackage[numbers]{natbib}
\usepackage{multicol}
\usepackage{amsmath} % assumes amsmath package installed
\usepackage{amssymb}  % assumes amsmath package installed
\usepackage{mathrsfs}

\usepackage[per-mode=symbol]{siunitx}
\DeclareSIUnit{\mph}{mph}
\usepackage{tikz}

\usepackage{pgfplots}
\pgfplotsset{compat=newest}
\pgfplotsset{every axis/.append style={
	font=\LARGE}
}
\pgfplotsset{every axis legend/.append style={legend cell align=left}}

\usetikzlibrary{shapes,arrows, automata}
\usetikzlibrary{arrows.meta, calc, shapes}
\tikzset{%
  >={Latex[width=2mm,length=2mm]},
            base/.style = {rectangle, rounded corners, draw=black,
                           minimum width=1cm, minimum height=1cm,
                           text centered, font=\sffamily},
            simulator/.style = {base, fill=green!30, minimum width=4cm},
            solver/.style = {base, fill=red!30},
            reward/.style = {base, minimum height=1.5cm},
            module/.style = {base, minimum width=2.5cm, minimum height=1.5cm, fill=blue!30},
            module2/.style = {base, minimum width=2.5cm, minimum height=1.5cm, fill=white},
            network/.style = {base, minimum width=2.5cm, minimum height=1.5cm, fill=white},
            state/.style = {base, minimum width=0.5cm, minimum height=1.0cm, fill=white},
            io/.style = {base, minimum width=0.5cm, minimum height=1.0cm, fill=white},
}

\usepackage{hyperref} 
\usepackage{cleveref}
\crefname{appsec}{Appendix}{Appendices}

\newcommand{\vect}[1]{\boldsymbol{\mathbf{#1}}}

\begin{document}

\title{The Adaptive Stress Testing Formulation}

% You will get a Paper-ID when submitting a pdf file to the conference system
\author{
\authorblockN{Mark Koren}
\authorblockA{mkoren@stanford.edu \vspace*{-0.55cm}} 
\and 
\authorblockN{Anthony Corso}
\authorblockA{acorso@stanford.edu \vspace*{-0.55cm}}
\and
\authorblockN{Mykel J. Kochenderfer}
\authorblockA{mykel@stanford.edu \vspace*{-0.55cm}}}

%\author{\authorblockN{Michael Shell}
%\authorblockA{School of Electrical and\\Computer Engineering\\
%Georgia Institute of Technology\\
%Atlanta, Georgia 30332--0250\\
%Email: mshell@ece.gatech.edu}
%\and
%\authorblockA{Twentieth Century Fox\\
%\and
%\authorblockN{James Kirk\\ and Montgomery Scott}
%\authorblockA{Starfleet Academy\\
%San Francisco, California 96678-2391\\

% avoiding spaces at the end of the author lines is not a problem with
% conference papers because we don't use \thanks or \IEEEmembership

%Homer Simpson\authorrefmark{2},
%Montgomery Scott\authorrefmark{3} and
%Eldon Tyrell\authorrefmark{4}}
%Telephone: (800) 555--1212, Fax: (888) 555--1212}
%\authorblockA{\authorrefmark{4}Tyrell Inc., 123 Replicant Street, Los Angeles, California 90210--4321}}

\maketitle

\begin{abstract}
Validation is a key challenge in the search for safe autonomy. Simulations are often either too simple to provide robust validation, or too complex to tractably compute. Therefore, approximate validation methods are needed to tractably find failures without unsafe simplifications. This paper presents the theory behind one such black-box approach: adaptive stress testing (AST). We also provide three examples of validation problems formulated to work with AST.
\end{abstract}

\IEEEpeerreviewmaketitle

\section{Introduction}
An open question when robots operate autonomously in uncertain, real-world environments is how to tractably validate that the agent will act safely. Autonomous robotic systems may be expected to interact with a number of other actors, including humans, while handling uncertainty in perception, prediction and control. Consequently, scenarios are often too high-dimensional to tractably simulate in an exhaustive manner. As such, a common approach is to simplify the scenario by constraining the number of non-agent actors and the range of actions they can take. However, simulating simplified scenarios may compromise safety by eliminating the complexity needed to find rare, but important failures. Instead, approximate validation methods are needed to elicit agent failures while maintaining the full complexity of the simulation.
% Safety validation is a key challenge in the search for safe autonomy. Many systems of interest are safety-critical; therefore field-testing may be prohibitively dangerous. Unfortunately, autonomous robotic systems may be expected to deal with a number of other actors, including humans, while dealing with uncertainty in both perception and prediction. Consequently, scenarios are often too high-dimensional to tractably simulate in an exhaustive manor. However, simulating simplified scenarios may compromise safety by eliminating the complexity needed to find rare, but important failures. As such, validation methods will need to be able to approximately search the space of possible scenarios to tractably identify failures in complicated autonomous systems.

One possible approach to approximate validation is adaptive stress testing (AST)~\cite{lee2015adaptive}. In AST, the validation problem is cast as a Markov decision process (MDP). A specific reward function structure is then used with reinforcement learning algorithms in order to identify the most-likely failure of a system in a scenario. Knowing the most-likely failure is useful for two reasons: 1) all other failures are at most as-likely, so it provides a bound on the likelihood of failures, and 2) it uncovers possible failure modes of an autonomous system so they can be addressed. AST is not a silver bullet: it requires accurate models of all actors in the scenario and is susceptible to local convergence. However, it allows failures to be identified tractably in simulation for complicated autonomous systems acting in high-dimensional spaces. This paper briefly presents the latest methodology for using AST and includes example validation scenarios formulated as AST problems. 

\section{Methodology}

\subsection{Adaptive Stress Testing}
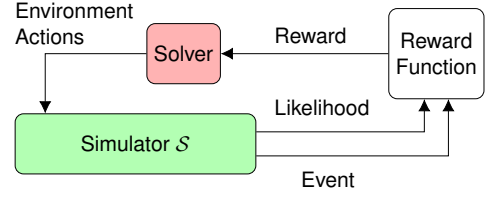
\begin{figure}[t]
    \setlength\belowcaptionskip{-0.75\baselineskip}	
	\centering
    \scalebox{0.8}{\begin{tikzpicture}[node distance=1.5cm,
    every node/.style={fill=white, font=\sffamily, text centered}, align=center]
  % Specification of nodes (position, etc.)
	\node (sim)             [simulator]              {Simulator $\mathcal{S}$};
    \node (solver)          [solver, above of = sim, xshift = 0.8cm]              {Solver};
    \node (reward)          [reward, above of = sim, xshift = 5cm]              {Reward\\Function};  
  % Specification of lines between nodes specified above
  % with aditional nodes for description 
  \draw[->]					(solver.west) -| node[text width=1cm, xshift = 0mm, yshift = 5mm, text centered, align=center] {Environment\\Actions} ($ (sim.north) - (15mm, 0) $);
  \draw[->] 	($(sim.east) + (0mm,2mm)$) -| node[text width=1.6cm, xshift = -17mm, yshift = 4mm] {Likelihood} ($ (reward.south) + (-2mm, 0mm) $);
  \draw[->] 	($(sim.east) + (0mm,-2mm)$) -| node[text width=1cm, xshift = -20mm, yshift = -4mm] {Event} ($ (reward.south) + (2mm, 0mm) $);
  \draw[->]		(reward.west) -- ++(0mm,0) -- node[text width=1cm, xshift = 0mm, yshift = 3mm] {Reward}(solver.east);
\end{tikzpicture}}
    \caption{The AST methodology. The simulator is treated as a black box. The solver optimizes a reward based on transition likelihood and whether an event has occurred.}
	\label{fig:ASTStruct} 
\end{figure}

Adaptive stress testing formulates the problem of finding the most-likely failure of a system as a Markov decision process (MDP)~\cite{DMU}. Reinforcement learning (RL) algorithms can then be applied to efficiently find a solution in simulation. The process is shown in \Cref{fig:ASTStruct}. An RL-based solver outputs \textit{Environment Actions}, which are the control input to the simulator. The simulator resolves the next time-step by executing the environment actions and then allowing the system-under-test (SUT) to act. The simulator returns the likelihood of the environment actions and whether an event of interest, such as a failure, has occurred. The reward function, covered in \Cref{sec:reward}, uses these to calculate the reward at each time-step. The solver uses these rewards to find the most-likely failure using reinforcement learning algorithms such as Monte Carlo tree search (MCTS)~\cite{UCT} or trust region policy optimization (TRPO)~\cite{schulman2015trust}.

\subsection{Problem Formulation}
Finding the most-likely failure of a system is a sequential decision-making problem. Given a simulator $\mathcal{S}$ and a subset of the state space $E$ where the events of interest (e.g. a collision) occur, we want to find the most-likely trajectory $s_0, \ldots, s_t$ that ends in our subset $E$. Given $(\mathcal{S}, E)$, the formal problem is
% \begin{equation*}
% \begin{aligned}
% & \underset{X}{\text{minimize}}
% & & \mathrm{trace}(X) \\
% & \text{subject to}
% & & X_{ij} = M_{ij}, \; (i,j) \in \Omega, \\
% &&& X \succeq 0.
% \end{aligned}
% \end{equation*}
\begin{equation*}
\begin{aligned}
& \underset{a_0, \ldots, a_t}{\text{maximize}}
& & P(s_0, a_0, \ldots,s_t, a_t) \\
& \text{subject to}
& & s_t \in E
\end{aligned}
\end{equation*}
where $P(s_0, a_0, \ldots,s_t, a_t)$ is the probability of a trajectory in simulator $\mathcal{S}$ and $s_t = f(a_t, s_{t-1})$.
% According to the Markov assumption, $s_{t+1}$ only depends on $s_t$ and $a_t$.
% find the trajectory  s.t. $s_t \in E$ and $\max\limits_{[a_0,...,a_t]}$P$([s_0, ..., s_t]|\mathcal{S})$.

% The process is shown in~\Cref{fig:ASTStruct}. The AST solver outputs an action (referred to as environment actions to differentiate them from the actions of the SUT), which is used by the simulator $\mathcal{S}$ to advance the environment forward a time-step. The simulation returns a transition probability and an indicator of whether the current state is terminal. The reward function uses the transition probability and terminal flag to pass a reward back to the solver. The solver then optimizes the policy, and then outputs the next environment action.

AST requires the following three functions to interact with the simulator:
\begin{itemize}
\item \textsc{Initialize}$(\mathcal{S}, s_0)$: Resets $\mathcal{S}$ to a given initial state $s_0$.
\item \textsc{Step}$(\mathcal{S}, E, a)$: Steps the simulation in time by drawing the next state $s'$ after taking action $a$. The function returns the probability of the transition and an indicator showing whether $s'$ is in $E$ or not.
\item \textsc{IsTerminal}$(\mathcal{S}, E)$: Returns true if the current state of the simulation is in $E$ or if the horizon of the simulation $T$ has been reached. 
\end{itemize}

\subsection{Reward Function} \label{sec:reward}
In order to find the most-likely failure, the reward function must be structured as follows:
\begin{equation}
\label{eq:1}
R\left(s\right) = \left\{
        \begin{array}{ll}
            0 &  s \in E \\
            -\alpha - \beta f(s) &  s \notin E, t\geq T \\
            -g(a) - \eta h(s)  &  s \notin E, t < T
        \end{array}
    \right.
\end{equation}
where the parameters are:
\begin{itemize}
    \item $\alpha$: A large number, to heavily penalize trajectories that do not end in the target set.
    \item $\beta f(s)$: An optional heuristic. For example, in the autonomous vehicle experiment, we use the distance between the pedestrian and the car at the end of a trajectory. Consequently, the network takes actions that move the pedestrian close to the car early in training, allowing collisions to be found more quickly.
    \item $g(a)$: The action reward. A function recommended to be  something proportional to $\log P(a)$. Adding log-probabilities is equivalent to multiplying probabilities and then taking the log, so this constraint ensures that summing the rewards from each time-step results in a total reward that is proportional to the log-probability of a trajectory. 
    % Optimizing to maximize reward results in the most likely failure trajectory.
    \item $\eta h(s)$: An optional training heuristic given at each timestep.
\end{itemize}

Looking at \Cref{eq:1}, there are three cases:
\begin{itemize}
    \item $s \in E$: The trajectory has terminated because an event has been found. This is the goal, so the reward at this step is as large as possible (\SI{0}{}).
    \item $s \notin E, t\geq T$: The trajectory has terminated by reaching the horizon $T$ without reaching an event. This is the least-useful outcome, so the user should set a large penalty.
    \item $s \notin E, t < T$: A time-step that was non-terminal, which is the most common case. The reward is generally proportional to the negative log-likelihood of the environment action, which promotes likely actions.
\end{itemize}

Ignoring heuristics for now, it is clear that the reward will be better for even a highly-unlikely trajectory that terminates in an event compared to a trajectory that fails to find an event. However, among trajectories that find an event, the more-likely trajectory will have a better reward. Consequently, optimizing to maximize reward will result in maximizing the probability of a trajectory that terminates with an event.

\section{Examples}

We present three scenarios in which an autonomous system needs to be validated. For each scenario, we provide an example of how it could be formulated as an AST problem. Further details available in \Cref{app:examples}.

\subsection{Cartpole with Disturbances}

\subsubsection{Problem}

Cartpole is a classic test environment for continuous control algorithms~\cite{cartpole}. The system under test (SUT) is a neural network control policy trained by TRPO. The control policy controls the horizontal force $\vec{F}$ applied to the cart, and the goal is to prevent the bar on top of the cart from falling over.

\subsubsection{Formulation}

We define an event as the pole reaching some maximum rotation or the cart reaching some maximum horizontal distance from the start position. The environment action is $\delta \vec{F}$, the disturbance force applied to the cart at each time-step. The reward function uses $\alpha = \SI{1e4}{}$, $\beta = \SI{1e3}{}$, and $f(s)$ as the normalized distance of the final state to failure states. The choice of $f(s)$ encourages the solver to push the SUT closer to failure. The action reward, $g(a)$ is set to the log of the probability density function of the natural disturbance force distribution. See \citet{ASTToolbox}.
% given by
% \begin{equation}
% f(s) = \min\left(\frac{|x-x_max|}{x_max},\frac{|\theta-\theta_max|}{\theta_max}\right)
% \end{equation}
\subsection{Autonomous Vehicle at a Crosswalk}

\subsubsection{Problem}

Autonomous vehicles must be able to safely interact with pedestrians. Consider an autonomous vehicle approaching a crosswalk on a neighborhood road. There is a single pedestrian who is free to move in any direction. The autonomous vehicle has imperfect sensors. 

\subsubsection{Formulation}

A collision between the car and pedestrian is the event we are looking for. The environment action vector controls both the motion of the pedestrian as well as the scale and direction of the sensor noise. The reward function for this scenario uses $\alpha = \SI{-1e5}{}$ and $\beta = \SI{-1e4}{}$, with $f(s) = \textsc{dist}\left(\vect p_v,\vect p_p\right)$ as the distance between the pedestrian and the SUT at the end of a trajectory. This heuristic encourages the solver to move the pedestrian closer to the car in early iterations, which can significantly increase training speeds. The reward function also uses $g(a) = M\left(a, \mu_a\mid s\right)$, which is the Mahalanobis distance function~\cite{mahalanobis1936generalised}. Mahalanobis distance is a generalization of distance to the mean for multivariate distributions. See \citet{koren2018}.

\subsection{Aircraft Collision Avoidance Software}

\subsubsection{Problem}
The next-generation Airborne Collision Avoidance System (ACASX)~\cite{kochenderfer2012next} gives instructions to pilots when multiple planes are approaching each other. We want to identify system failures in simulation to ensure the system is robust enough to replace the Traffic Alert and Collision Avoidance System (TCAS)~\cite{kuchar2007traffic}. We are interested in a number of different scenarios in which two or three planes are in the same airspace.

\subsubsection{Formulation}
The event will be a near mid-air collision (NMAC), which is when two planes pass within 100 vertical feet and 500 horizontal feet of each other. The simulator is quite complicated, involving sensor, aircraft, and pilot models. Instead of trying to control everything explicitly, our environment actions will output seeds to the random number generators in the simulator. The reward function for this scenario uses $\alpha =\infty$ and no heuristics. The reward function also uses $g(a) = \log P(s_t \mid s_{t+1})$, the log of the known transition probability at each time-step. See \citet{lee2015adaptive}.

\section{Conclusion}
This paper presents the latest formulation of adaptive stress testing, and examples of how it can be applied. AST is an approach to validation that can tractably find failures in autonomous systems in simulation without reducing scenario complexity. Autonomous systems are difficult to validate because they interact with many other actors in high-dimensional spaces according to complicated policies. However, validation is essential for producing autonomous systems that are safe, robust, and reliable. 

\bibliographystyle{plainnat}
\bibliography{references}

\clearpage
% \newpage
% \pagebreak
% \newpage
% \appendix
\begin{appendices}
    \crefalias{section}{appsec}
    \section{Examples: Further Details} \label{app:examples}
    \subsection{Cartpole with Disturbances}
\begin{figure}[h]
    \centering
    % \vspace*{0.5cm}
    \includegraphics[width=0.5\columnwidth,height=0.5\columnwidth]{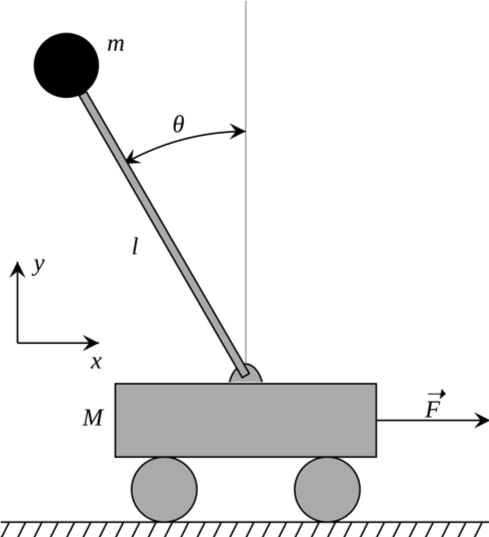}
    \caption{Layout of the cartpole environment. A control policy applies horizontal force on the cart to prevent the bar falling over.}%\todohere{too big / make structure of captions match}}
    \label{fig:cartpole}
    \end{figure}

    The cartpole scenario from \citet{ASTToolbox} is shown in \Cref{fig:cartpole}. The state $s=[x,\dot{x},\theta,\dot{\theta}]$ represents the cart's horizontal position and speed as well as the bar's angle and angular velocity. The control policy, a neural network trained by TRPO, controls the horizontal force $\vec{F}$ applied to the cart. The failure of the system is defined as $|x|>x_{max}$ or $|\theta|>\theta_{max}$. The initial state is at $s_0=[0,0,0,0]$.
    
    \subsection{Autonomous Vehicle at a Crosswalk}
\begin{figure}[htbp]
	
	\centering
    % \vspace*{0.25cm}
    \centering
    \includegraphics[width=0.70\columnwidth]{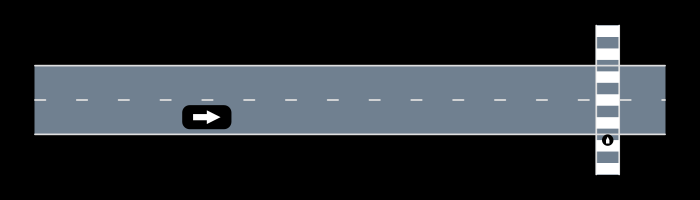}
    \caption{Layout of the autonomous vehicle scenario. A vehicle approaches a cross-walk on a neighborhood road as a single pedestrian attempts to walk across.}
	\label{fig:scenario1} 
    \end{figure}

    The autonomous vehicle scenario from \citet{koren2018} is shown in \Cref{fig:scenario1}. The $x$-axis is aligned with the edge of the road, with East being the positive $x$-direction. The $y$-axis is aligned with the center of the cross-walk, with North being the positive $y$-direction. The pedestrian is crossing from South to North. The vehicle starts \SI{35}{\meter} from the crosswalk, with an initial velocity of \SI{11.2}{\meter\per\second} East. The pedestrian starts \SI{2}{\meter} away, with an initial velocity of \SI{1}{\meter\per\second} North. The autonomous vehicle policy is a modified version of the intelligent driver model~\cite{PhysRevE621805}.
    
    \subsection{Aircraft Collision Avoidance Software}
\begin{figure}[h]
	
	\centering
    % \vspace*{0.25cm}
    \centering
    \includegraphics[width=0.70\columnwidth]{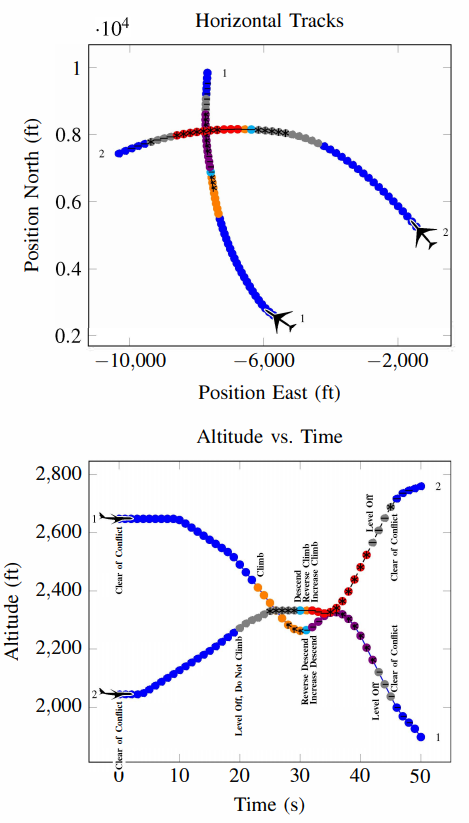}
    \caption{An example result from \citet{lee2015adaptive}, showing an NMAC identified by AST. Note that the planes must be both vertically and horizontally near to each other to register as an NMAC.}
	\label{fig:lee} 
    \end{figure}

    An example result from \citet{lee2015adaptive} is shown in \Cref{fig:lee}. The planes need to cross paths, and the validation method was able to find a rollout where pilot responses to the ACASX system lead to an NMAC. AST was used to find a variety of different failures in ACASX. 

\end{appendices}

\end{document}